# A Novel Temporal Multi-Gate Mixture-of-Experts Approach for Vehicle Trajectory and Driving Intention Prediction

**Renteng Yuan (First author)**
Jiangsu Key Laboratory of Urban ITS
School of Transportation
Southeast University, Nanjing, Jiangsu, P. R. China, and 210000
Email: rtengyuan123@126.com

**Mohamed Abdel-Aty**
Department of Civil, Environmental and Construction Engineering
University of Central Florida, 12800 Pegasus Dr #211, Orlando, FL 32816, USA
Email: M.aty@ucf.edu

**Qiaojun Xiang (Corresponding author)**
Jiangsu Key Laboratory of Urban ITS
School of Transportation
Southeast University, Nanjing, Jiangsu, P. R. China, and 210000
Email: xqj@seu.edu.cn

**Zijin Wang**
Department of Civil, Environmental and Construction Engineering
University of Central Florida, 12800 Pegasus Dr #211, Orlando, FL 32816, USA
Email: zijinwang@knights.ucf.edu

**Ou Zheng**
Department of Civil, Environmental and Construction Engineering
University of Central Florida, 12800 Pegasus Dr #211, Orlando, FL 32816, USA
Email: ouzheng1993@knights.ucf.edu

**ABSTRACT**
Accurate Vehicle Trajectory Prediction is critical for automated vehicles and advanced driver assistance systems. Vehicle trajectory prediction consists of two essential tasks, i.e., longitudinal position prediction and lateral position prediction. There is a significant correlation between driving intentions and vehicle motion. In existing work, the three tasks are often conducted separately without considering the relationships between the longitudinal position, lateral position, and driving intention. In this paper, we propose a novel Temporal Multi-Gate Mixture-of-Experts (TMMOE) model for simultaneously predicting the vehicle trajectory and driving intention. The proposed model consists of three layers: a shared layer, an expert layer, and a fully connected layer. In the model, the shared layer utilizes Temporal Convolutional Networks (TCN) to extract temporal features. Then the expert layer is built to identify different information according to the three tasks. Moreover, the fully connected layer is used to integrate and export prediction results. To achieve better performance, uncertainty algorithm is used to construct the multi-task loss function. Finally, the publicly available CitySim dataset validates the TMMOE model, demonstrating superior performance compared to the LSTM model, achieving the highest classification and regression results.

**INTRODUCTION**

Vehicle intelligence, automation, and cooperative control are transformative trends in transportation. Vehicle trajectory prediction plays a critical role in various fields, such as autonomous driving, advanced driver assistance systems (ADAS), and intelligent transportation systems. Accurately predicting vehicle trajectory can help intelligent in-vehicle devices understand the surrounding traffic environment, identify potential safety risks, and formulate effective control strategies.

Vehicle trajectories represent the accumulation of driving behavior over time and space. It can be represented by longitudinal and lateral position changes. Predicting the vehicle trajectory is not a deterministic task since it is influenced by vehicle dynamics, driver maneuvers, and the surrounding traffic environment *(1, 2)*. Additionally, changes in driving intentions can directly impact vehicle trajectories. Studies indicated that changes in driving intentions are a significant factor contributing to errors in vehicle trajectory prediction *(3-5)*. In previous research, driving intention recognition modules have been utilized prior to vehicle trajectory prediction *(1, 6)*. While improving accuracy, these modeling frameworks face the challenge of building separate predictive models for each metric (e.g., intention, longitudinal and lateral position), which is limited to some particular scenarios (e.g., lane changing process)*(7)*. It is also recognized that there exists a correlation between the longitudinal and lateral position of vehicles *(8)*. Considering the correlation between longitudinal and lateral positions as well as driving intentions, it is necessary to develop a multi-task prediction model for accurate vehicle trajectory prediction.

The multi-task learning (MTL) model, initially proposed by Caruana in 1997 *(9)*, is a specialized type of deep learning model designed to simultaneously train and learn multiple related tasks. As a promising area in machine learning, MTL aims to improve the performance of multiple related learning tasks by leveraging useful information across tasks *(10)*. This paper presents a novel methodology, referred to as the Temporal Multi-Gate Mixture-of-Experts (TMMOE) model, for simultaneously recognizing driving intention and predicting vehicle trajectory. The proposed model comprises three layers: a shared layer, an expert layer, and a fully connected layer (FCN). The shared layer utilizes Temporal Convolutional Networks (TCN) to extract temporal features from the input data. The expert layer employs Long Short-Term Memory (LSTM) to adaptively differentiate between tasks and capture long-term dependencies in the time series data. Each task is associated with an FCN layer, which utilizes the extracted features to generate target values.

**LITERATURE REVIEW**

There are three commonly employed methods for predicting vehicle trajectories: physics-based models, maneuver-based models, and learning-based models *(4, 11)*. Physics-based methods represent vehicles as dynamic entities governed by the laws of physics. These models focus on establishing vehicle motion model or utilizing fitting techniques to model vehicle trajectories *(12-14)*. Common methods include polynomials *(15, 16)*, Gaussian processes *(17)*, and Kalman filter *(13)*. However, these methods do not consider interactions between the surrounding vehicles in vehicle trajectory prediction and are limited to short-term motion prediction *(7, 18)*. Maneuver-based methods involve early recognition of driving intention and classify vehicle trajectories into a finite set of typical patterns *(1)*. Markov models (HMMs), Bayesian networks, and Support vector machine (SVM) classifiers are three commonly used methods *(6, 14, 19)*. Nevertheless, these methods classify vehicle motion into distinct categories, ignoring the continuous and holistic nature of driver behavior. In recent years, learning-based models, particularly deep learning methods, have gained significant attention. One such method is the recurrent neural network (RNN), which utilizes large-scale datasets to learn complex relationships and improve the accuracy of vehicle trajectory prediction *(20)*.

The Long Short-Term Memory (LSTM) Network is a special type of Recurrent Neural Network (RNN) that has been widely used in the prediction of vehicle trajectories *(21, 22)*. In order to enhance the predictive performance, several studies have incorporated an attention mechanism into LSTM models *(11, 14, 23)*. However, there are still two limitations that have not been addressed in these approaches: the issue of vanishing gradients and the lack of parallel computation capability *(24)*. To overcome these limitations, Zeng et al.*(25)*, Islam et al. *(26)*, and Xie et al.*(27)* proposed a CNN-LSTM sequential model for vehicle trajectory prediction. In their work, the convolutional neural network (CNN) was employed to extract data features and input them into the LSTM network. However, CNN employs fixed-size filters to handle input data as independent samples, limiting its ability to effectively model temporal correlations and sequential structures. Bai et al.*(28)* designed a special CNN, Temporal Convolutional Network (TCN), for processing sequential data, such as time series or natural language *(29-31)*. With dilated causal convolution layers, the TCN can effectively capture long-term dependencies across multiple time scales within input sequences. Previous research reported that the TCN model has achieved significant promotion in both regression and classification tasks *(31-34)*.

Multi-tasks learning is a special deep learning framework. In traditional multi-task learning, a neural network is typically utilized to simultaneously predict multiple tasks. Previous research has shown that the intricate relationships among tasks have a significant impact on the performance of multi-task learning models *(10, 35, 36)*. In practical engineering applications, the correlation among tasks may dynamically change with time or spatial dimensions. For instance, vehicle states during lane changing are more sensitive to the surrounding traffic environment than the car following process *(8)*. The Multi-gate Mixture-of-Experts (MMOE) model, initially proposed by Ma et al.*(37)*, is a novel approach in multi-task learning that effectively handle task differences without requiring explicit task difference measurements. The MMOE model combines expert and gate networks to capture task relationships among tasks and learn task-specific functionalities. It has achieved significant promotion in both regression and classification tasks, involving health status *(38)*, energy systems load prediction *(39)*, and sound classification *(40)*. While the MMOE (Mixture of Experts) model is developed within the framework of the DNN (Deep Neural Network) structure, further advancements are needed to optimize its effectiveness in handling time series data.

As mentioned above, vehicle trajectory prediction involves simultaneously predicting multiple indicators. To enhance the accuracy of predicting vehicle trajectories, this study proposes an improvement to the MMOE model. It incorporates TCN networks for feature extraction and LSTM networks for gating memory mechanisms. This integration enhances the effectiveness of the model in handling time series data and improves the accuracy of trajectory prediction.

**METHOD**

The objective of this study is to develop a unified model that can simultaneously recognize driving intentions and predict vehicle trajectories. Vehicle trajectories are represented by both the lateral and longitudinal positions. The driving intentions are classified into three categories: lane keeping (LK), lane changing to the left (LCL), and lane changing to the right (LCR). In order to achieve this, the model had to learn three tasks simultaneously, which include two regression tasks and one classification task. It is very challenging to directly observe and quantify the complex mapping relationships between changes in longitudinal and lateral positions. To address these challenges in a unified model, this research proposed a novel data-driven multi-task learning (MTL) model for the recognition of driving intentions and the prediction of vehicle trajectories.

**The Proposed Model**

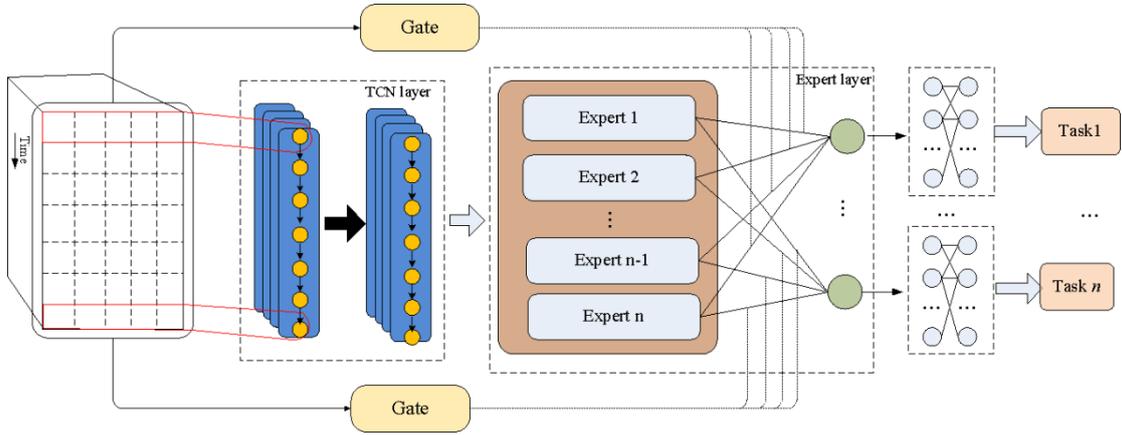

**Figure 1 Flowchart of the proposed model**

As shown in Figure 1, the proposed model includes a shared layer, an expert layer, and a fully connected (FCN) layer. Based on the input data, the shared layer is employed to extract temporal features. The Expert layer employs a multi-gate mixture-of-experts (MMoE) network structures to adaptively differentiate tasks and extract valuable information. The shared information extracted from the shared layer is separately used as the input of the expert network. The gate network helps the model learn task-specific information for each task. The aggregated results from the experts are subsequently fed into the task-specific FCN layer. Each specific task is associated with an FCN layer, which leverages the obtained features to generate the target values.

The following introduces three layers: the shared layer, the expert layer, the FCN layer. Moreover, the joint loss function for model training is described.

**The Shared Layer**

The Temporal Convolutional Network (TCN) is utilized in the shared layer to extract temporal features. TCN is composed of causal convolution and dilated convolution *(28, 29)*. Causal convolutions are used to ensure the temporal dependencies of the input data. Dilated convolutions are incorporated into causal convolutions to increase the receptive field, allowing the model to capture a wider range of temporal information. The one-dimensional fully-convolutional network (1DFCN) architecture is employed to produce the same length output as the input *(41)*. For a filter $f: \{1, 2, \ldots, k-1\}$, the dilated convolution operation F on the element $s$ of a 1-D sequence $x \in R^n$ is formulated as,

$$F(s) = (x *_d f)(s) = \sum_{i=0}^{k-1} f(k) \cdot x_{s - d \cdot i} \quad (1)$$

Where $d$ is the dilation parameter and is used to control the size of the interval, $k$ is the filter size and represents the number of convolution kernels, * is the convolution operator, $s - d \cdot i$ accounts for the direction of the past. The dilated causal convolution structure is depicted in Figure 2.

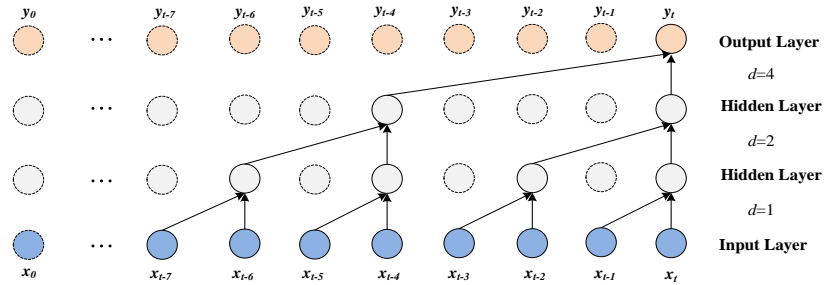

**Figure 2** A dilated causal convolution with dilation factors d = 1,2,4 and **kernel** size k = 2 *(42)*

As shown in Figure 2, the kernel size is set to 2, and the depth of the causal convolution is 3. The convolution indicated that the output at time t is associated with the input points from time t-7 to time t. Residual blocks are used to address disappearance and gradient expansion in TCN. By incorporating techniques such as longer convolutional kernels and residual connections, TCN is able to capture long-term dependencies effectively. Figure 3 illustrates the architecture of the residual block, utilizing a Rectified Linear Unit (ReLU) as the activation function and incorporating batch normalization for the convolutional filter. A 1x1 convolution is added in the residual block when the input and output data have different lengths.

Yuan, Abdel-Aty, Xiang, Wang, and Zheng

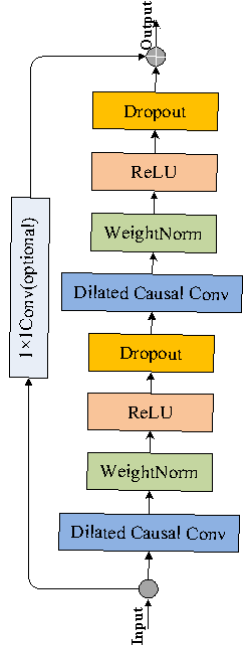

**Figure 3 TCN residual block** *(8)*

**The Expert Layer**

In this section, we present notable enhancements to the MMOE model, with a specific emphasis on its utilization for time series forecasting and analysis. The output of task k can be described as:

$$y_t^k = h^k\left(f^k\left(x_t\right)\right) \qquad (2)$$

$$f^k\left(x_t\right) = \sum_{i=1}^{n} g^k\left(x_t\right)_i f_i\left(x_t\right) \qquad (3)$$

Where $x_t$ represents the input sequence at time *t*, $y_t^k$ represents the output indicator at time *t* for task k. $f^k$ represents the shared bottom network task k, and $h^k$ represent the tower networks where k =1, 2, ..., K for each task respectively. $f_i(x_t)$ is the *ith* expert network. $g^k(x_t)_i$ represents the gating network for task k, and is the weight for the *i th* expert network, and $\sum_{i=1}^{n} g^k(x_t)_i = 1$. In our proposed model, the gating networks are implemented using LSTM cells and then ensemble the results from all experts. The weights of the expert networks are dynamically adjusted based on the input time series. The architecture of the gating networks can be described as follows.

$$g\left(x_t\right) = softmax\left(h\left(x_t\right)\right) \qquad (4)$$

$$h\left(x_t\right) = o_t \odot tanh\left(c_t\right) \qquad (5)$$

$$c_t = f_t \odot c_{t-1} + i_t \odot \tilde{c}_t \qquad (6)$$

Where $c_t$ is the memory cell at time *t-1*, $\tilde{c}_t$ is the candidate memory at time t, *h(x_t)* is the outcome at time t. ⊙ represents vector element-wise product. $i_t$ is the input gate, $f_t$ is the forget gate, and $o_t$ is the output gate. These gates are computed as follows.

$$i_t = \sigma(W_i x_t + U_i h_{t-1} + b_i) \quad (7)$$

$$f_t = \sigma(W_f x_t + U_f h_{t-1} + b_f) \quad (8)$$

$$o_t = \sigma(W_o x_t + U_o h_{t-1} + b_o) \quad (9)$$

Where $\sigma$ represents the sigmoid activation function; $x_t$ represents the input sequence at time $t$; $h_{t-1}$ represents the hidden state; $W$ is the parameter matrix at time $t$ and represents the input weight; U is the parameter matrix at time $t$ -$1$, and represents the recurrent weight; $b_i$, $b_f$, and $b_o$ represents bias. In addition, each expert network is built as an independent LSTM model. The internal update state of the LSTM recurrent cells can be expressed as:

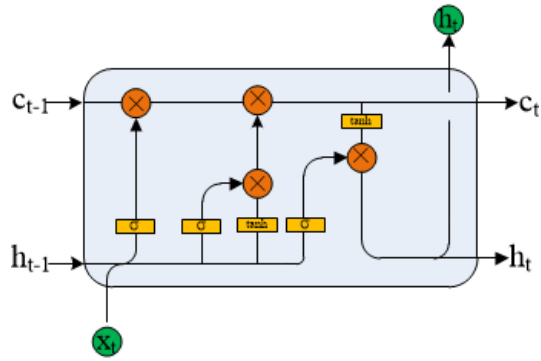

**Figure 4 LSTM block structure**

**The FCN Layer**

The FCN layer, serving as the final output module in our proposed model, consists of independent network layer structures. The results of the assembled experts are then passed to the task-specific FCN networks. The FCN layer leverages the aggregated information to perform task-specific processing and produce the final outputs. Each task is associated with its own dedicated FCN layer. The Rectified Linear Unit (ReLU) activation function was selected as the activation function.

**Multi-task Loss Function Based on Homoscedastic Uncertainty**

This research involves two types of tasks: classification and regression. In this paper, we utilize the cross-entropy loss function to achieve driving intention classification. Cross-entropy is a commonly used loss function that effectively measures the difference between predicted probabilities and true labels *(43)*. It offers advantages in gradient computation and is based on the concept of information theory. This loss functions *L(θ)* are defined as follows:

$$L_c(\theta) = -\frac{1}{N} \sum_i \sum_{c=1}^{M} y_{ic} \log(p_{ic}) \quad (10)$$

Where *θ* represents the model parameters, M denotes the number of classes, $y_{ic}$ represents a sign function where y equals 1 if the true class of sample *i* is equal to c, and 0 otherwise, $p_{ic}$ represents the predicted probability that observation i belongs to class c.

Vehicle trajectory prediction is a regression issue. The Mean-Squared Loss is a prevalent loss function employed in regression tasks. It quantifies the average squared difference between the predicted values and the corresponding true values of the target variable. Minimizing the Mean-Squared Loss enables models to optimize their parameters, aiming to achieve the best possible fit to the training data and reduce overall prediction errors. In this research, the Mean-Squared Loss is adopted as the loss function for the regression task. The formula for Mean-Squared Loss is as follows:

$$L_r(\theta) = \frac{1}{N}\sum_{i=1}^{N}(y_i - \hat{y}_i)^2 \quad (11)$$

Where N represents the number of samples, $y_i$ represents the *i-th* sample labels, and $\hat{y}_i$ represents the predicted value for the *i-th* sample.

The weight settings of the joint loss functions across tasks have a significant influence on the performance of the model. The homoscedastic uncertainty method facilitates a balanced optimization process for learning diverse quantities with varying units or scales through a tradeoff in the joint loss functions*(44)*. Considering the inherent uncertainty of different tasks, this paper utilizes the homoscedastic uncertainty method to dynamically adjust the weights of the joint loss functions. For classification tasks, the softmax function is employed to estimate probabilities. The probability function and log likelihood for the classification task can be expressed as follows.

$$p(y|f_c^\omega(x), \sigma_c) = \text{Softmax}\left(\frac{1}{\sigma_c^2} f_c^\omega(x)\right) \quad (12)$$

$$\log p(y = n|f_c^\omega(x), \sigma_c) = \frac{1}{\sigma_c^2} f_{c=n}^\omega(x) - \log \sum_{n'} \exp\left(\frac{1}{\sigma_c^2} f_{c=n'}^\omega(x)\right) \quad (13)$$

Where $f_c^\omega(x)$ represents the prediction result of a neural network with weight ω on input x. $y$ is the corresponding label. $\sigma_c^2$ is a positive scaling factor, which can be interpreted as a Gibbs distribution where the input is scaled by $\sigma_c^2$. $f_{c=n'}^\omega(x)$ represents the $n'$th element of the vector $f_c^\omega(x)$. For regression tasks, the likelihood is defined as a Gaussian with mean given by the model output. The log likelihood for this output can be obtained by substituting the Gaussian distribution function into the log-likelihood function, resulting in the following expression.

$$p(y|f_r^\omega(x), \sigma_r) = N(f_r^\omega(x), \sigma_r^2) \quad (14)$$

$$\log N(y|f_r^\omega(x), \sigma_r) = -\log \sigma_r - \frac{1}{2\sigma_c^2}\|y - f_r^\omega(x)\|^2 \quad (15)$$

In our proposed model, the outputs comprise three vectors $f_c^\omega(x)$, $f_{r_1}^\omega(x)$ and $f_{r_2}^\omega(x)$. These vectors are modeled using two regression likelihoods and one classification likelihood. So the multi-task loss function $L$ (ω, $\sigma_c$, $\sigma_{r1}$, $\sigma_{r2}$) is given as

$$L(\theta, \sigma_c, \sigma_{r_1}, \sigma_{r_2}) \approx \frac{1}{\sigma_c^2} L_c(\theta) + \frac{1}{2\sigma_{r_1}^2} L_{r_1}(\theta) + \frac{1}{2\sigma_{r_2}^2} L_{r_2}(\theta) + \log\sigma_c + \log\sigma_{r_1} + \log\sigma_{r_2} \quad (16)$$

The detailed derivation process can be found in Yan et al*(43)*, Kendall et al *(44)*, and Liu et al*(45)*. Algorithm 1 shows the method for determining the optimal parameters in the proposed model based on the homoscedastic uncertainty function.

---

**Algorithm 1** The proposed model optimization by using the homoscedastic uncertainty function.

---

**Input:**
Training feature x, classification label $y_c$ and regression label $y_{r_1}, y_{r_2}$; the maximum epoch $E_{\max}$, batch size $\vartheta$; Model initial parameters $\theta$ and loss function parameters $\sigma_c$, $\sigma_{r_1}$ and $\sigma_{r_2}$.

**Output:** $\theta$, $\sigma_c$, $\sigma_{r_1}$ and $\sigma_{r_2}$.

1: i←0 ;∕ ∗The index of the epoch.∗ ∕
2: **while** $i < E_{\max}$ **do**
3: **for** each batch $\vartheta$ **do**
4: Input feature data $x$, classification label $y_c$ and regression label $y_{r_1}, y_{r_2}$ into the model;
5: Update model parameters $\theta, \sigma_c, \sigma_{r_1}$ and $\sigma_{r_2}$ using Adam optimizer to find the minimum of $L(\theta, \sigma_c, \sigma_{r_1}, \sigma_{r_2}) = \frac{1}{\sigma_c^2}L_c(\theta) + \frac{1}{2\sigma_{r_1}^2}L_{r_1}(\theta) + \frac{1}{2\sigma_{r_2}^2}L_{r_2}(\theta) + log\sigma_c + log\sigma_{r_1} + log\sigma_{r_2}$;
6: **end for**
7: i←i+1
8: **end while**
9: Return the $\theta, \sigma_c, \sigma_{r_1}, \sigma_{r_2}$ and stop

---

## DATASET

Experimental data were obtained from the publicly available CitySim dataset *(46)*. The CitySim dataset is a collection of vehicle trajectory data obtained from drones. So far it includes data from 14 locations and has a high sampling frequency of 30 Hz. Notably, the dataset supplies the vehicle center, head, tail, and bounding box vertices locations, thereby enabling an intricate assessment of their movements. We chose a sub-dataset, freeway-B, with six lanes in two directions to evaluate the performance of our proposed model. Figure 5 shows a snapshot of the freeway-B segment. The freeway-B dataset was collected using two UAVs simultaneously, covering a 680 meters basic freeway segment. Each lane has a width of 3.75 meters. A total of 5623 vehicle trajectories were extracted from 60 minutes of drone videos.

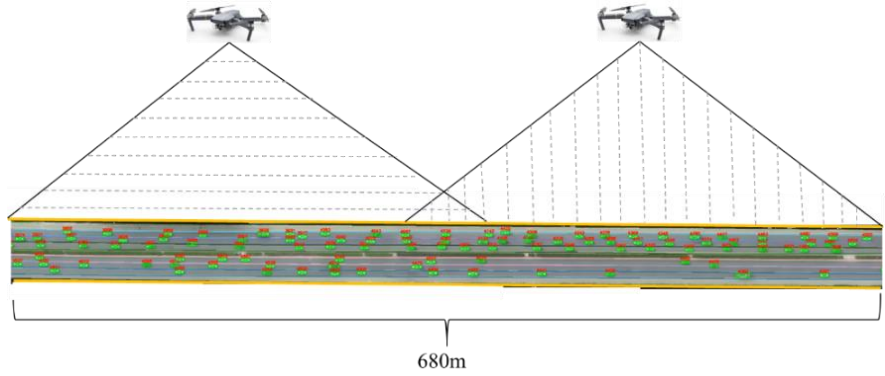

**Figure 5 A snapshot of the freeway-B segment**

**Data Processing**

In the original dataset, the position of each lane marking was represented by extracting 8 to 10 points, manually selected from the provided images. It is important to acknowledge that there may be inherent errors or uncertainties in the accuracy of these points. To mitigate these issues, this study leverages available information, such as the lane width and the positional coordinates of the points, to reconstruct the position information of the lane markings. The reconstruction process is depicted in Figure 6.

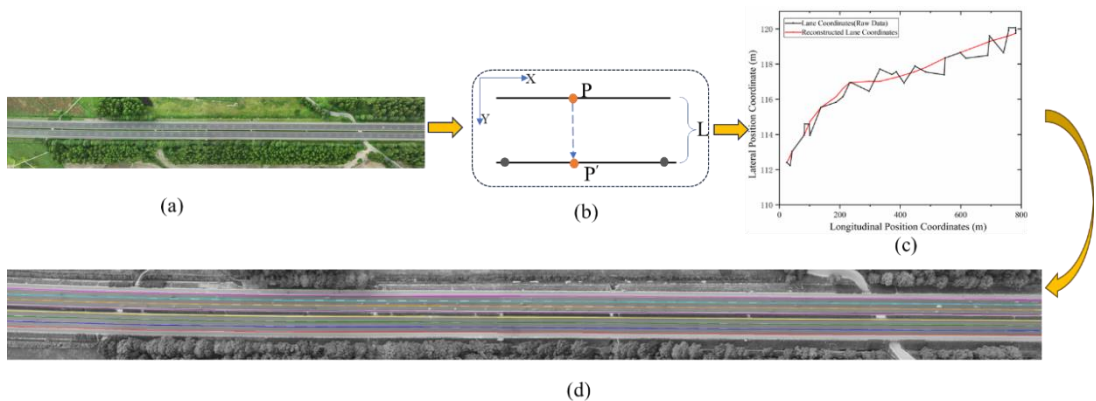

Figure 6(a) depicts the actual lane markings captured in the drone video.
Figure 6(b) depicts the projection operation applied to the lane markings.
Figure 6(c) illustrates the reconstructed lane markings after the projection process.
Figure 6(d) shows the overlay of the reconstructed lane markings onto the original image.
**Figure 6 Reconstructed Lane Coordinates**

To begin with, all the points on the lane markings are projected onto the outermost lane line by subtracting the corresponding lane width. Following this, a moving average algorithm is employed to process the projected coordinate values, as depicted in the Figure 6(c). Next, a translation operation is conducted to determine the coordinates for the remaining lane lines. It is crucial to acknowledge that even minor positioning errors can significantly affect the extraction indicators, such as speed and acceleration *(47)*. To mitigate the negative impact of these errors, a moving average (MA) method is utilized to smooth the trajectory. The moving average filter is set to 15 frames (corresponding to a duration of 0.5 seconds), and the sampling rate is configured at 3. A comparative analysis of the original trajectory and the processed trajectory is depicted in Figure 7. For a more in-depth explanation of the data processing methods used in this study, please refer to our previous study *(8)*.

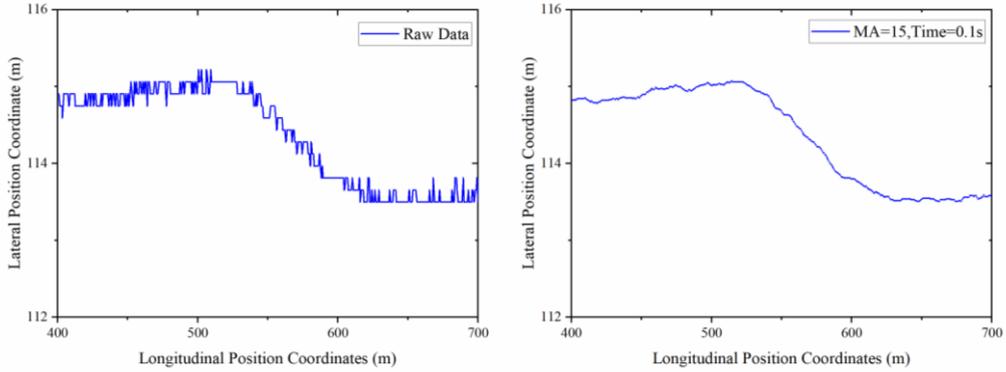

**Figure 7 Comparison of original trajectory and processed trajectory**

In total, 1047 vehicle trajectories were extracted from the freeway-B dataset. This includes 647 lane change (LC) vehicle trajectories, comprising 272 left lane change (LCL) trajectories and 375 right lane change (RCL) trajectories. Additionally, 400 lane-keeping (LK) vehicle trajectories were randomly selected from the dataset. To mitigate the numerical differences' influence on network training, the lateral position is adjusted accordingly. In the CitySim dataset, the local coordinate y represents the distance from the survey image boundary. With a known road width of 3.75m, all vehicle trajectory origins are shifted to the middle lane. The extracted partial vehicle trajectory is shown in Figure 8. Figure 8(a) represents the raw vehicle trajectory. Figure 8(b) depicts the processed vehicle trajectory.

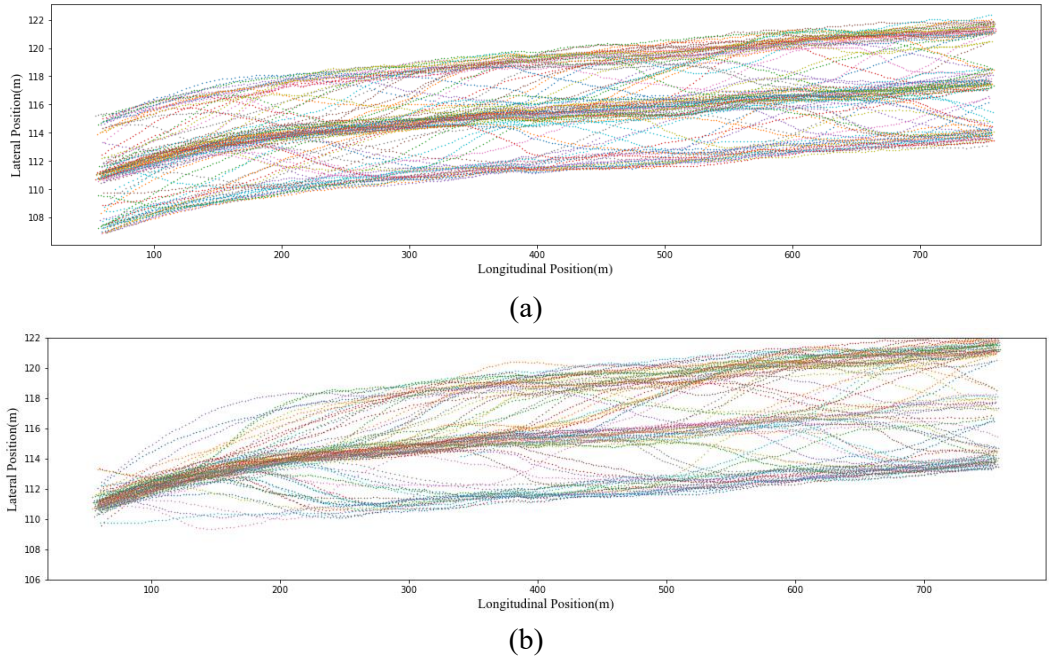

**Figure 8 the extracted partial vehicle trajectory**

**Driving Intention Labeling**

Within this study, the driving intentions are classified into three distinct categories: lane keeping (LK), left lane change (LCL), and right lane change (RCL). Each sequence extracted from the vehicle trajectories can be represented as either LCL, RCL, or LK. The lane-

Yuan, Abdel-Aty, Xiang, Wang, and Zheng

changing process consists of two stages: the lane change decision (LCD) and the lane change execution (LCE). In this study, the LCE is defined as the period when the vehicle's front bounding box points touch the target lane boundary until the vehicle's rear bounding box points enter the target lane. Figure 9 shows one event, RCL, where point D touches the lane boundary and is recognized as the start time of an LCE process.

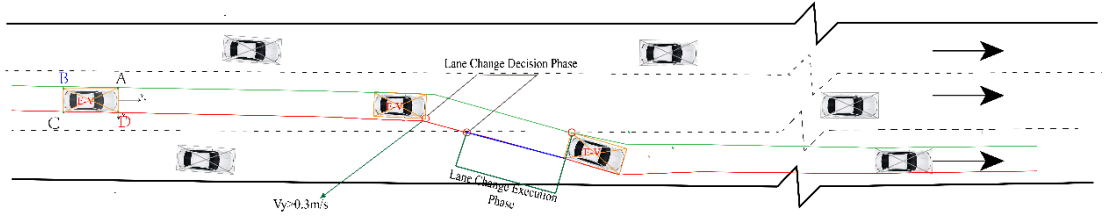

**Figure 9 Lane changing process division**

The LCD occurs before the execution of the lane change. In general, the initiation of LCD is defined as the point in time when the lateral velocity of the vehicle surpasses 0.1 m/s (or -0.1m/s) for the first time *(48, 49)*. Based on Figure 6(c), it can be observed that the lane line exhibits an approximate lateral deviation of 8m. Additionally, based on statistical analysis, it has been observed that vehicles travel for an average duration of approximately 26 seconds on this road segment. Consequently, in this study, the LCD is determined based on the lateral velocity ($v_y$) exceeding 0.3m/s, which indicates the LCD to the right. Conversely, the LCD to the left is determined when the lateral velocity ($v_y$) falls below 0m/s.

The driver's response time ranges from 0.3 to 1.35 seconds, while the braking system's activation time is approximately 0.15 seconds *(50)*. This study reveals that vehicles have an average duration of 1.70 seconds for the LCD process, while the average duration for the LCE is 2.71 seconds. To capture the process of lane change execution (LCE), this study sets the predicted time window to 3 seconds. During the labeling process, the extracted sequences are categorized based on the location of their endpoints. If the endpoint falls within the lane change decision (LCD) stage, the sequence is labeled as either left lane change (LLC) or right lane change (RLC). On the other hand, if the endpoint is outside the LCD stage, the sequence is labeled as lane keeping (LK). Furthermore, the data labels are encoded as (0, 1, 0) for LLC, (0, 0, 1) for RLC, and (1, 0, 0) for LK

**EXPERIMENT AND RESULTS**

This section focuses on comparing the proposed model with existing models and the subsequent quantitative analysis of the results. The dataset is randomly split into a training dataset and test dataset with a ratio of 8:2. A ten-fold cross-validation method was used for model training and evaluation on the training dataset. All models are trained within the Keras 2.11.0 framework. The experimental operations are conducted on an Intel(R) Core (TM) i7-4770 CPU @3.40GHz hardware environment.

**Input Variables**

The movement of the target vehicle is influenced by other vehicles. The input variables in this research are divided into three categories: vehicle trajectory information and vehicle interaction information. The vehicle trajectory information comprises trajectories of the target vehicle and surrounding vehicles, including the nearest preceding and following vehicles in

*Yuan, Abdel-Aty, Xiang, Wang, and Zheng*

adjacent and current lanes. Vehicle interaction information is used to describe the interaction between vehicles. To evaluate the driving status of vehicles, we first introduce the concept of system coupling coordination degree. Each vehicle is treated as an independent subsystem, while the target vehicle and its adjacent surrounding vehicles constitute a complete system. The system coupling coordination degree is calculated using the instantaneous speeds of each vehicle, as shown in the following expression.

$$C_t = \frac{n\sqrt[n]{v_{1,t} v_{2,t} v_{3,t} \ldots v_{n,t}}}{v_{1,t}+v_{2,t}+v_{3,t}+..+v_{n,t}} \quad (17)$$

Where $C_t$ represents the coordination degree within the system at time t for a given target vehicle. $v_{i,t}$ represents the instantaneous speed of vehicle i at time t. *n* represents the number of vehicles, including the target vehicle and surrounding vehicles. The maximum value of n is 7. If the surrounding corresponding vehicle does not exist, the corresponding *v* value is not included in the formula. A lower $C_t$ value indicates greater speed differentials among vehicles. Three traffic conflict indicators were used to measure the collision risk of target vehicles, including TTC (time to collision)*(51)*, MTTC (modified time-to-collision) *(52, 53)*, and DRAC (Deceleration Rate to Avoid a Crash) *(54, 55)*. The conflict values between this vehicle and the front and rear vehicles in the target lane, as well as the front and rear vehicles in the same lane, are extracted. The conflict indicator is defined as follows.

$$\text{Conflict indicator} = \begin{cases} 0 & (\text{have no conflict}) \text{ if } MTTC = TTC = DRAC = 0 \\ 1 & (\text{have conflict}) \text{ otherwise} \end{cases} \quad (18)$$

Where TTC, MTTC, and DRAC are all binary variables with values of 0 and 1, where 1 indicates the presence of a conflict, and 0 indicates no conflict. Both the TTC threshold and MTTC threshold are set to 2.5 seconds, and the DRAC threshold is set to 3.35 m/s², based on previous studies *(52, 56)*.

**Experimental Details**
The performance of a model is inherently affected by its parameter settings. In order to identify the optimal parameter configuration, we conducted sensitivity experiments using the control variable method on four different models. This research chose the parameters based on two key metrics: classification accuracy and training time. Our goal was to find the optimal parameter settings that strike a balance between these two factors. Specifically, we aim to minimize the training time and reduce the complexity of the model while ensuring that the accuracy of the model remains uncompromised. In the shared layer, the size of dilated convolution interval is set to $\{2^1, 2^2, 2^3, \ldots, 2^n\}$, which depends on the input time series length, along with 64 filters, one stack of residual blocks, and a kernel size of 2. In the expert layer, the number of expert networks is set to 12, and each expert network is implemented using an independent LSTM model with 64 units. The FCN layer for each task consists of 64 units, respectively, and the ReLU is used as the activation function. The batch size is set to 256, and training epochs are set to 200. The optimizer is Adam.

**Performance Evaluation Metrics of Intention and Trajectory Prediction**
To measure the performance of the model, classification tasks were assessed using the

Yuan, Abdel-Aty, Xiang, Wang, and Zheng

area under the curve (AUC) and accuracy metrics, while regression tasks were evaluated based on the Root Mean Squared Error (RMSE) and Mean Absolute Error (MAE). To investigate the effect of input sequence length on classification outcomes, samples with different data lengths D (D = 3s, 6s, 9s) are extracted, before the LCE process. In each epoch, the loss function is computed independently for the test set. As an example, Figure 10 depicts the convergence process of the loss function during the training of the proposed model with 6s as the input length.

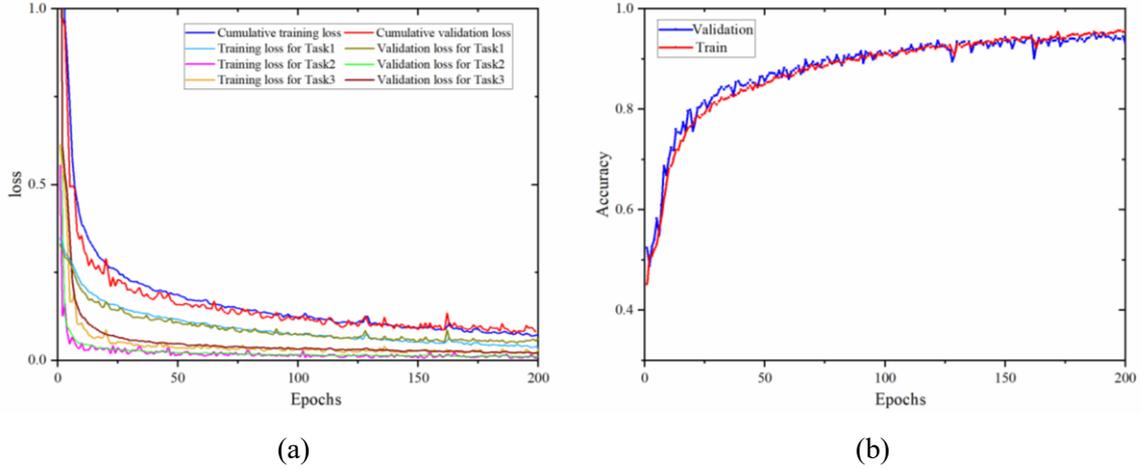

(a)                  (b)

**Figure 10 Convergence process of loss function during training with 6s as input.**

In Figure 10(a), the cumulative loss is obtained by utilizing the loss function of homoscedastic uncertainty. Task 1 represents driving intention classification, Task 2 represents longitudinal position prediction, and Task 3 represents lateral position prediction. Figure 10(b) presents the classification accuracy of driving intention. Figure 10 demonstrates the convergence of the prediction model, as four loss functions consistently decrease over time, proving the proposed model's convergence. To compare the performance of the proposed model, we also employ the LSTM model with different input data lengths, and the results are summarized in Table 1.

Table 1 Model performance comparison

| Model | Task | 3s | | | 6s | | | 9s | | |
|---|---|---|---|---|---|---|---|---|---|---|
| | | Acc | RMSE | MAE | Acc | RMSE | MAE | Acc | RMSE | MAE |
| TMMOE | Inte | 0.9222 | -- | -- | 0.9544 | -- | -- | 0.9123 | -- | -- |
| | Lon | -- | 2.5756 | 1.878 | -- | 2.323 | 1.785 | - | 3.081 | 2.197 |
| | Lat | -- | 0.3067 | 0.2039 | -- | 0.2963 | 0.199 | - | 0.415 | 0.362 |
| LSTM | Inte | 0.8471 | -- | -- | 0.8796 | -- | -- | 0.8931 | -- | |
| | Lon | -- | 7.831 | 5.8726 | -- | 8.280 | 6.1573 | -- | 9.723 | 7.3337 |
| | Lat | -- | 0.646 | 0.4768 | -- | 0.6274 | 0.4839 | -- | 0.645 | 0.4785 |

**Note**: Inte represents driving intention; Lon represents longitudinal position; Lat represents lateral position; Acc represents average accuracy.

Table 1 shows that the performance (regression and classification) of the model proposed in this study outperforms the corresponding LSTM model for a specific input length. The TMMOE model achieved the highest performance in the classification and regression tasks when the input time series was set to 6 seconds. Specifically, table 1 presents that the TMMOE model showed a significant improvement in classification accuracy compared to LSTM models, increasing from 0.9222 to 0.9544 with a six-second input time series. Hence, a time duration of six seconds was chosen as the input sequence length. To comprehensively evaluate the performance of the TMMOE model, the Receiver Operating Characteristic (ROC)curve for each class is shown in Figure 11.

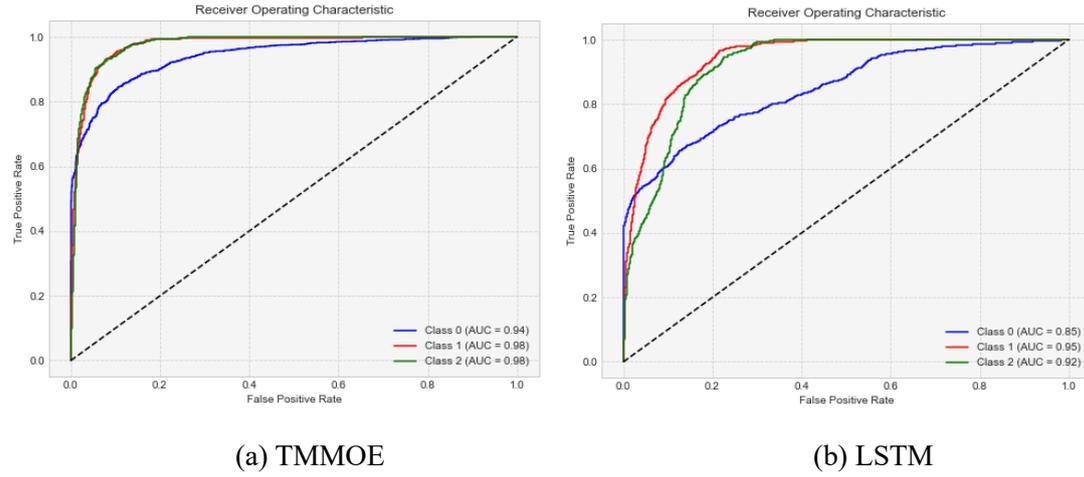

(a) TMMOE  (b) LSTM

**Figure 11 The ROC curve of driving intention classification with input 6 seconds**

In Figure 11, the class labels are defined as follows: Class 0 represents the LK samples, Class 1 represents the RLC, and Class 2 represents the LLC samples. Figure 11(a) shows the ROC curve for TMMOE, and Figure 11(b) shows the ROC curve for LSTM. As shown in Figure 11(a), the AUC for Class 0 is 0.94, while the AUC for both Class 1 and 2 is 0.98. The results indicate that the proposed model achieved higher AUC values for each class compared to LSTM. It can be concluded that the TMMOE model has a strong ability to effectively differentiate between the three classes. In addition, it is worth noting that the misidentification of LK categories as RLC and LLC is identified as the primary source of classification errors.

In terms of vehicle position prediction results, the TMMOE model shows promising vehicle position prediction results with an RMSE of 2.323m and an MAE of 1.785m for the longitudinal position, as well as an RMSE of 0.2963m and a MAE of 0.199m for the lateral position. The TMMOE model demonstrates significant improvement over the LSTM model, for instance, reducing the RMSE from 8.280m to 2.323 for longitudinal prediction and from 0.6274m to 0.2963m for lateral prediction. In addition, the RMSE index is significantly lower compared to previous studies *(14, 18)*. The results show that considering the correlation between driving intentions, longitudinal and lateral positions are crucial for accurately predicting vehicle trajectories. Figure 12 displays the longitudinal position prediction results for selected samples. Figure 12 illustrates the lateral position and trajectory prediction results.

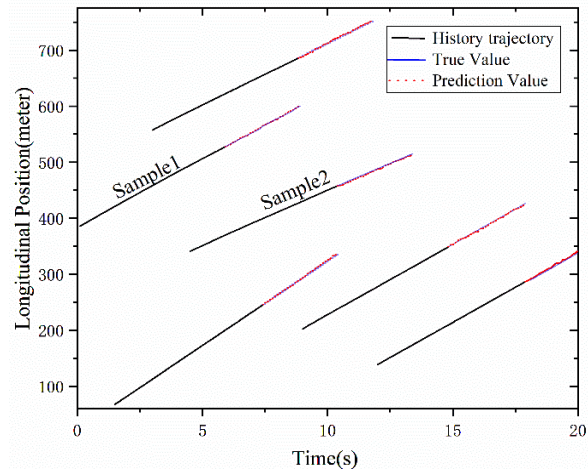

**Figure 12 Vehicle longitudinal position prediction results**

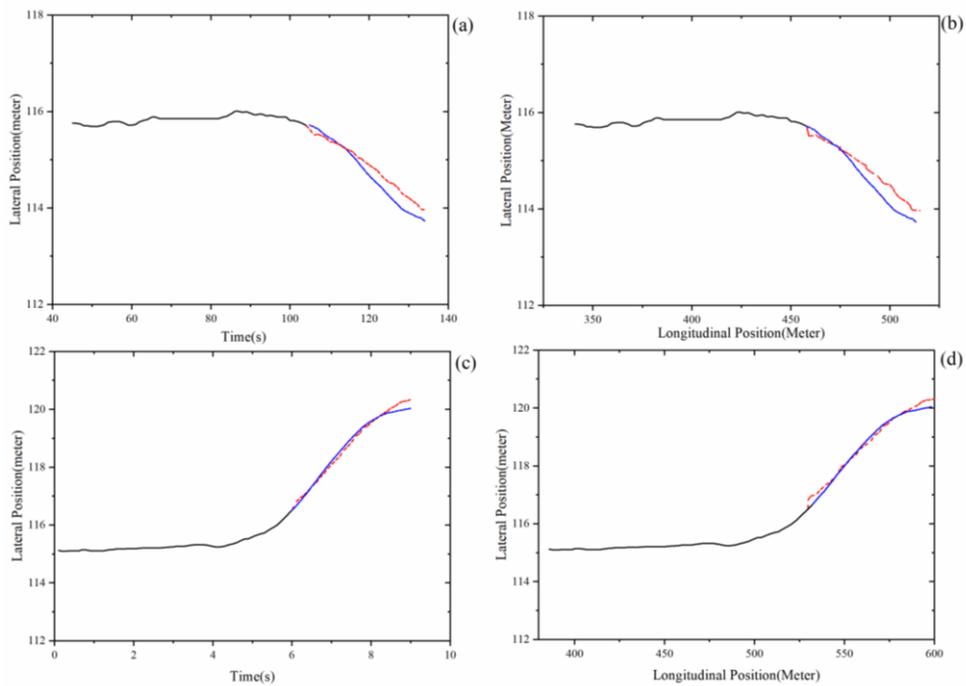

**Figure 13 Vehicle lateral position prediction results**

Figure 13(a) and 13(c) depict the vehicle's lateral position prediction results over time. Figure 13(b) and 13(d) illustrate the vehicle trajectory prediction results at a spatial scale. Figure 13(a) and 13(b) correspond to sample 1 from Figure 12. Figures 13(c) and 12(d) correspond to sample 2 from Figure 12.

**Sensitivity analysis of Decision Variables**

Sensitivity analysis was conducted to verify the rationality of the conflict indicator and coupling degree used in the proposed TMMOE model. The control variable method is a commonly used approach in sensitivity analysis *(57)*. The testing procedure involved removing one of the input variables (i.e., conflict indicator and coupling degree) at a time to create a new vehicle trajectory prediction model. The importance of each variable was then analyzed by comparing the changes in predicted outcomes when the variable was removed from the model. The evaluation metrics of the TMMOE model with one decision variable

*Yuan, Abdel-Aty, Xiang, Wang, and Zheng*removed are presented in Table 2. It is evident that removing the conflict indicator and coupling degree variable resulted in a decrease in classification accuracy and an increase in the RMSE and MAE. Specifically, the accuracy decreased from 0.9544 to 0.9157 and 0.8907, respectively, compared to the original TMMOE model. The results highlight the significance of traffic conflict indicator and coupling degree variable in predicting vehicle trajectories.

Table 2 Evaluation metrics of the TMMOE model when one of the decision variables is removed.

| Remove variable | Task | Acc | RMSE | MAE |
| --- | --- | --- | --- | --- |
| Coupling degree | Inte | 0.9157 | -- | -- |
| | Lon | -- | 2.4021 | 2.848 |
| | Lat | -- | 0.2888 | 0.201 |
| Conflict indicator | Inte | 0.8907 | -- | -- |
| | Lon | -- | 3.1512 | 2.592 |
| | Lat | -- | 0.3013 | 0.2041 |

**Note**: Inte represents driving intention; Lon represents longitudinal position; Lat represents lateral position; Acc represents average accuracy.

## CONCLUSIONS

In this research, we construct a novel Temporal Multi-Gate Mixture-of-Experts (TMMOE) model for simultaneously predicting the vehicle trajectory and driving intention. The proposed model is capable of simultaneously recognizing the driving intention and predicting the vehicle's trajectory. The architecture of the proposed approach consists of a shared layer, an expert layer, and a fully connected layer. In order to capture temporal features effectively, the TCN was integrated into the shared layer. Furthermore, by utilizing LSTM units as the gating mechanism, the expert layer is able to effectively model and retain relevant information over extended sequences, thereby enhancing its capacity for long-term dependency modeling. To achieve better performance, homoscedastic uncertainty algorithm is used to construct the multi-task loss function. The publicly available CitySim dataset is used to validate the performance of the TMMOE model. Nonetheless, there are some shortcomings and limitations of this study. Only one dataset, CITYSIM, was used in this study. Therefore, future studies should utilize additional datasets to enhance the validation of the model's portability. This study exclusively focuses on evaluating the model's performance using input time series lengths of 3, 6, and 9 seconds. However, it could be worth attempting to incorporate finer input time series lengths.

## ACKNOWLEDGEMENT

This work was supported in part by the Postgraduate Research & Practice innovation Program of Jiangsu Province (No.KYCX22_0270), and the China Scholarship Council (CSC).## AUTHOR CONTRIBUTIONS

The authors confirm contribution to the paper as follows: study conception and design, draft manuscript preparation: RentengYuan; Review & Editing: Mohamed Abdel-Aty; Academic guidance and advice: Mohamed Abdel-Aty, QiaojunXiang; Suggestion,data collection and processing: OuZheng, Zijin Wang, RentengYuan; All authors reviewed the results and approved the final version of the manuscript.